\documentclass[twocolumn]{article}
\usepackage[dvipdfmx]{graphicx}
\usepackage{arxiv}
\usepackage{authblk}

\usepackage{amsmath,amssymb,amsfonts,amsthm}
\usepackage{mathrsfs}%
\usepackage[utf8]{inputenc}
\usepackage{hyperref}
\usepackage{enumitem}

\usepackage{xcolor}%
\usepackage{textcomp}%
\usepackage{manyfoot}%
\usepackage{algorithm}%
\usepackage{algorithmicx}%
\usepackage{algpseudocode}%
\usepackage{listings}%

\usepackage{multirow}
\usepackage{subfig}
\usepackage{array}
\usepackage{booktabs}
\usepackage{threeparttable}
\usepackage{colortbl}

\theoremstyle{thmstyleone}%
\theoremstyle{thmstyletwo}%

\theoremstyle{thmstylethree}%
\newcommand{\ccol}[2]{ \multicolumn{#1}{c}{#2}}

\newcolumntype{P}[1]{>{\centering\arraybackslash}m{#1}}

\begin{document}

\newcommand{\myPaperShortTitle}{Hierarchical Memorization in LLMs}
\newcommand{\myPaperTitle}{Hierarchical Memorization in Large Language Models: Evidence from Citation Generation}
\title{\myPaperTitle}
\date{}

%\author{Junichiro Niimi}

\renewcommand\Authfont{\bfseries}
\setlength{\affilsep}{0em}
% box is needed for correct spacing with authblk
\newbox{\orcid}\sbox{\orcid}{\includegraphics[scale=0.06]{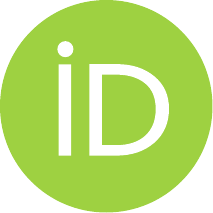}} 
\author[1]{%
	\href{https://orcid.org/0000-0002-4618-6272}{\usebox{\orcid}\hspace{1mm}
	Junichiro Niimi\thanks{\texttt{jniimi@meijo-u.ac.jp}}
	}}
\affil[1]{Meijo University}

\renewcommand{\shorttitle}{\myPaperShortTitle}
\newcommand{\FOne}{macro\text{-}F_1}
\newcommand{\RMSE}{R\hspace{-0.05em}M\hspace{-0.1em}S\hspace{-0.07em}E}

%\maketitle

\twocolumn[
	\begin{@twocolumnfalse}
		\maketitle
\vspace{-3em}
\begin{abstract}
Large language models (LLMs) generate fluent text across a wide range of tasks, but the fabrication of non-existent academic citations remains a critical and well-documented failure mode. Building on prior work that frames hallucination and verbatim memorization as outcomes of the same probabilistic process, this study uses citation count as a proxy for training data redundancy and asks how this redundancy is internally structured within a single bibliographic record. Using GPT-4.1, we generated and manually verified 100 citations across twenty computer-science domains, measuring factual fidelity via cosine similarity against authentic metadata. We find that (i) factual accuracy varies substantially across domains and scales log-linearly with citation count, (ii) the model crosses two empirically identifiable thresholds; an inflection around 90 citations and a saturation point near 1,200 citations beyond which records are reproduced nearly verbatim, (iii) memorization is hierarchical, with titles and first authors recalled earliest while venues and numeric fields require far greater redundancy and publication years remain essentially unlearned, and (iv) even highly cited records can be conflated when their titles and authors overlap, an effect interpretable as spurious-attractor interference. Memorization in LLMs is therefore not a binary on/off state but a graduated, hierarchically layered phenomenon shaped by the uneven distribution of knowledge in the pretraining corpus.
\end{abstract}
\vspace{0.5em}
\keywords{Natural Language Processing \and Large Language Models \and Recommendation \and Hallucination \and Citation}
\vspace{2em}
	\end{@twocolumnfalse}
]

\renewcommand\thefootnote{*}
\setcounter{footnote}{0}

\newcommand{\bccol}[2]{ \multicolumn{#1}{c}{\bfseries{#2}}}

\section{Introduction}
\subsection{Background}
Large language models (LLMs) have achieved remarkable fluency across a wide range of domains \cite{llm_sentiment_review}. 
However, they are also known to generate hallucinations that are nonsensical or unfaithful to the provided source content \cite{hallucination,hallucination_inevitable1}. 
In particular, the generation of non-existent academic references or legal precedents has been widely recognized as a critical issue \cite{hallucination2}. 
For example, in the field of marketing, where Recency–Frequency–Monetary (RFM) analysis \cite{rfm1,rfm2,rfm3} is commonly employed as a customer relationship management (CRM) \cite{loyalty}, when prompted to “Please suggest recent academic papers on RFM analysis with Author (Year) Title, Journal, Vol, No, pp style,” the model (GPT-4.1) produced the following response:
\begin{quote}
Chitturi, P., Raghunathan, B., Sciandra, R., \& Sikora, J. (2010). 
“RFM and CLV: Using Customer Data for Improved Decision Making.” 
Journal of Direct, Data, and Digital Marketing Practice, 12(1), 1--10.
\end{quote}
Although the output follows the correct bibliographic format, the paper itself does not exist. The record appears to be composed of elements from different genuine studies: some author names (e.g., Chitturi and Raghunathan \cite{hallu2}), the paper title (e.g., “RFM and CLV:” \cite{hallu1}), and the journal name \cite{hallu3}. The remaining part of the title and numerical details are fictitious, suggesting that multiple authentic entries were probabilistically combined into a fabricated citation.

Recent studies on training data memorization \cite{carlini2} have shown that the probability of reproducing training data correlates with its frequency of appearance in pretraining corpora. Given that corpora include various web sources such as academic publications, technical blogs, and online discussions, highly cited papers which frequently appear across such sources are likely to be more accurately reproduced.

This study therefore focuses on citation recommendation using LLMs and empirically examines how factual correctness varies with citation prominence. We hypothesize that citation count serves as a proxy for training data redundancy (i.e., the frequency with which a given bibliographic record appears in the pretraining corpus). The experiment is conducted in a \textit{closed system}, where LLMs must rely solely on pretraining knowledge without access to external retrieval tools or databases. While retrieval-augmented approaches can mitigate hallucination by grounding outputs in external sources, understanding how LLMs behave when operating purely from internal knowledge remains critical; both for characterizing the boundaries of memorization and for scenarios where external retrieval is impractical or unavailable. 

\subsection{Contributions}
Our preliminary conference paper \cite{niimi_icaart} provided initial evidence for the citation--memorization relationship by analyzing 100 bibliographic records across 20 computer science domains with rigorous manual verification, and proposed that hallucination and memorization arise not randomly but systematically from imbalanced knowledge distributions in the pretraining corpus, as two sides of the same probabilistic process. Building on this view, the present extended study examines the same process in greater depth and finds that the transition between fabrication and faithful recall is not a clean dichotomy but a graduated, hierarchical phenomenon, in which different metadata fields within a single bibliographic record cross the memorization threshold at different citation levels. We therefore reframe the central claim from a binary contrast between hallucination and memorization toward a layered account in which memorization itself is internally structured.

Concretely, while the preliminary work established the basic correlation between citation frequency and factual accuracy and identified memorization thresholds, the present version substantially extends that work in three ways: (i) a new domain-level analysis of factual accuracy across research areas (Experiment~1), (ii) a quantitative field-level analysis demonstrating hierarchical memorization patterns across seven metadata fields in Section~5.1, and (iii) a theoretical interpretation of memory interference drawing inspiration from the framework of modern Hopfield network \cite{ramsauer2021} in Section~5.2. In addition, the related work has been expanded to cover LLM fundamentals, training data memorization, and recent studies on citation hallucination.

The remainder of this paper is organized as follows: Section 2 reviews related study on hallucination and memorization in LLMs and citation recommendation. Section 3 describes our experimental design. Section 4 presents experimental analyses of the relationship between memorization and pretraining data redundancy. Section 5 analyzes error patterns, focusing on hierarchical memorization and memory interference. Finally, Section 6 concludes with implications and limitations.

\section{Related Work}
\subsection{Large Language Models}
LLMs are neural language models pretrained on large-scale corpora using autoregressive next-token prediction, where the model learns to predict each successive token given the preceding context \cite{zhao2023}. The pretraining corpora typically comprise diverse web sources, including web pages, books, academic publications, technical blogs, and online discussions \cite{gpt3,llama}. For instance, the LLaMA \cite{llama} training corpus explicitly includes arXiv papers and StackExchange posts alongside Common Crawl web data. Through this process, LLMs acquire broad linguistic and factual knowledge, which is implicitly encoded in their parameters, often referred to as parametric knowledge \cite{LLM_as_KB}.

However, the distribution of information within these corpora is inherently uneven. Popular topics, frequently discussed entities, and highly cited publications appear far more often than niche or recent content. Since autoregressive models learn by approximating the statistical distribution of their training data, this imbalance directly affects what the model ``knows'': information that appears frequently is reinforced and retained, while infrequent information may be only partially learned or absent entirely. This uneven distribution has two important consequences: hallucination and memorization, which we discuss in the following subsections.

\subsection{Hallucination in LLMs}
Hallucination in LLMs refers to the generation of content that is nonsensical or unfaithful to factual sources \cite{hallucination,hallucination_inevitable1}. This phenomenon has been examined from diverse perspectives, including natural language generation \cite{hallucination,hallucination2,hallucination4}, code generation \cite{hallucination_codegen}, and knowledge-intensive tasks \cite{hallucination3}.

A fundamental cause of hallucination lies in the training objective itself. Since LLMs are trained to predict the next token based on statistical patterns in the training corpus, they inherently favor fluency and plausibility over factual accuracy \cite{hallucination4}. Furthermore, reinforcement learning with human feedback (RLHF) \cite{rlhf,reinforce1} can exacerbate this tendency, as models are penalized for responding “I don’t know” (IDK) and instead rewarded for producing confident, well-formed responses \cite{openai_hallucinate}. This alignment objective can thus promote plausible but unreliable statements, particularly when the model’s parametric knowledge is insufficient to produce a correct answer.

In the domain of academic citations, the severity of hallucination has been quantitatively demonstrated: Walters and Wilder \cite{walters_fabrication} reported that 55\% of GPT-3.5-generated citations and 18\% of GPT-4-generated citations were entirely fabricated. More alarmingly, a systematic investigation of all papers published at major NLP conferences (ACL, NAACL, and EMNLP in 2024--2025) revealed that nearly 300 accepted papers contain at least one hallucinated citation, with the problem rapidly increasing over time \cite{hallucitation}. Furthermore, LLMs reflect human citation patterns but with a heightened bias toward highly cited papers \cite{algaba_citation_bias}, suggesting that LLM-generated references systematically amplify existing inequalities in scholarly visibility.

\subsection{Training Data Memorization}
While hallucination represents one failure mode of LLMs, security-oriented studies have highlighted the opposite tendency: information repeated multiple times during pretraining is more likely to be memorized and reproduced verbatim \cite{carlini2,carlini1,llm_leakage2,llm_leakage1}.
This view aligns with recent theoretical accounts positioning LLMs as probabilistic pattern recognizers that approximate data distributions rather than explicitly “understanding” knowledge \cite{kallens2024,mirchandani2023}. Kandpal et al. \cite{kandpal2023} provided further evidence that question-answering accuracy scales with the frequency of relevant documents in the pretraining corpus, demonstrating that this frequency--accuracy relationship extends beyond verbatim reproduction to broader knowledge retrieval tasks. 

From this perspective, hallucination and training data leakage represent opposite outcomes of the same probabilistic learning process, where the frequency of appearance in training data governs whether information is accurately recalled or probabilistically fabricated.

In the context of citation recommendation, this implies that frequently cited papers which appear across numerous publications and other web sources are more likely to be recalled verbatim by LLMs, whereas sparsely represented works tend to be fabricated. Thus, LLM-based citation recommendation can be understood as retrieval from memorized training data, where recommendation quality depends on the strength of memorization rather than algorithmic ranking. This study therefore hypothesizes that hallucination in citation recommendation is systematically related to {\it the training data redundancy} ({\it i.e.,} the frequency with which a given bibliographic record appears in pretraining corpora): Highly cited papers are expected to be more robustly represented, leading to lower hallucination rates, while limited-redundancy papers are more prone to plausible but non-existent references.

\subsection{Citation Recommendation}
Citation recommendation has evolved through various methods \cite{biblio_dnn_survey1,biblio_survey_classic2,biblio_survey_classic1,biblio_survey1,biblio_survey2}, which can be broadly categorized into several approaches: content-based filtering, machine learning methods such as collaborative filtering \cite{cf}, deep 
neural network (DNN) approaches such as Transformer \cite{transformer} and BERT \cite{bert}, and more recently, LLM approaches.

Early content-based systems \cite{biblio_citeseer} relied on document similarity measures such as TF-IDF \cite{tfidf} and citation co-occurrence patterns. Subsequently, collaborative filtering methods \cite{biblio_cf1,biblio_cf2} emerged, which incorporate multiple approaches such as Latent Dirichlet Allocation (LDA) \cite{lda} and Singular Value Decomposition (SVD) \cite{svd}. They utilize user-item interaction patterns to identify relevant publications through topic modeling and matrix factorization approaches. With the advancement of deep learning, multiple studies have proposed DNN-based citation recommendation models \cite{biblio_dnn_survey1}. Representative approaches in this category include Transformer-based \cite{biblio_dnn2} and BERT-based \cite{biblio_dnn1} models. 

More recently, LLMs have been explored for enhancing citation recommendation through advanced embedding techniques \cite{biblio_llm1}. However, these approaches rely on pre-existing bibliographic databases and utilize LLMs solely for encoding abstracts into embeddings, rather than for generative tasks. While this avoids hallucination issues inherent in text generation, it can only recommend articles within pre-constructed citation networks.

As LLMs with larger context windows continue to emerge, generative approaches that consider bibliographic metadata and detailed content of papers as contextual information may become feasible, potentially overcoming these database dependency limitations. However, such generative citation recommendation operates as a \textit{closed system}, where LLMs must rely solely on knowledge acquired during pretraining without access to external databases or retrieval tools, and inevitably faces the challenge of hallucination.

\section{Experimental Design}
We conducted four experiments with the following settings:

\paragraph{Model settings.}
In this study, we employ GPT-4.1 (accessed via API; knowledge cutoff: June 2024) to generate bibliographic records. To ensure comparability of citation counts across domains, we limit our investigation to computer science and select twenty actively studied topics (e.g., transformer \cite{transformer}, diffusion model \cite{diffusion}, retrieval-augmented generation \cite{rag}).

\paragraph{Prompt.} The prompt used to generate bibliographic information is shown in Fig.~\ref{fig:prompt}. To ensure structured output, we include a JSON schema and explicitly instruct the model to provide no additional explanation.

\begin{figure}[htb]
\begin{center}
\begin{lstlisting}
### Instruction:
You are an academic assistant that outputs structured bibliographic data in JSON format.
Please suggest 5 recent academic papers related to "{domain}".

Each paper should be represented as a JSON object following this schema:
{
  "author": "Author name(s) in APA style, e.g., 'Smith, J. & Tanaka, K.'",
  "year": 2023,
  "title": "Title of the paper",
  "journal": "Name of the academic journal",
  "volume": "12",
  "number": "3",
  "pages": "123--145"
}

Output must be a single valid JSON array of objects and contain **no additional explanation**.
If you are unsure about any field, please leave it as an empty string ("").
\end{lstlisting}
\caption{Prompt to generate bibliographic information \cite{niimi_icaart}}\label{fig:prompt}
\end{center}
\end{figure}

\paragraph{Sample size.} 
While a larger sample would be desirable, our preliminary experiments revealed two practical constraints: (i) simultaneously requesting many recommendations strains the model's effective context window, leading to an increased rate of hallucinated entries, and (ii) the JSON output format becomes inconsistent or malformed. To balance manual verification feasibility with sample size, we prompt the model to recommend five papers per topic, yielding 100 samples total (20 topics $\times$ 5 papers).

\paragraph{Data source.}
We used Google Scholar to verify the existence of generated papers and collect citation counts as of October 2025. Google Scholar provides broader coverage than Web of Science or Scopus by indexing preprint repositories (e.g., arXiv), conference proceedings, and technical reports. This coverage is essential for capturing recent influential work such as Llama 3 \cite{llama3}, which may not yet appear in traditional citation databases. Moreover, because Google Scholar aggregates citations from a wide range of web-accessible sources, including educational materials, technical blogs, and online discussions, its citation counts more closely approximate a paper's overall presence across the types of web content that constitute LLM pretraining corpora, rather than reflecting purely academic impact.

\paragraph{Human evaluation.}
We employ two complementary evaluation methods. First, each record is manually scored for factual accuracy. The existence of a referenced paper is confirmed primarily through title matching; minor inconsistencies in author or journal names alone do not preclude a match. Records are classified as completely correct (score = 2), partially hallucinated (score = 1; the record matches an existing paper but some metadata, such as author names, journal, or year, are inaccurate), or completely hallucinated (score = 0; the record does not match any known papers).

\paragraph{Computational evaluation.}
Second, for records confirmed to exist (score $>$ 0), we compute semantic similarity between generated and authentic metadata. While manual scoring captures existence of the paper, cosine similarity quantifies the degree of factual fidelity in a continuous scale, which is critical for analyzing the relationship with citation frequency. Similarity is calculated using Sentence-BERT embeddings \cite{sbert} (all-MiniLM-L6-v2), which effectively capture semantic correspondence in short text fragments such as bibliographic records. All text was converted to lowercase before embedding to ensure consistent comparison.

\section{Experimental Analysis}
We conducted four experiments using 100 generated bibliographic records. Before discussing the individual experiments, Table~\ref{tab:records} shows three representative examples illustrating the scoring scheme. The completely correct record achieves cosine similarity of 0.995, while the partially hallucinated record still attains 0.798 despite fabricated co-authors, journal name, volume, and page numbers. This indicates that even relatively high similarity scores (e.g., 0.798) can still reflect partial hallucination, and near-perfect similarity (close to 1.0) is necessary to confirm successful reproduction.

\begin{table*}[htb]
   \caption{Examples of generated bibliographic records with varying degrees of hallucination \cite{niimi_icaart}. Fabricated elements are shown in bold. Cosine similarity is computed between generated and authentic metadata.}\label{tab:records}
   \scalebox{1}{
   \begin{tabular}{
m{16cm}
   }
\toprule
\bccol{1}{Generated results}\\
\midrule
Liu, Z., Lin, Y., Cao, Y., Hu, H., Wei, Y., Zhang, Z., Lin, S., \& Guo, B. (2021). Swin Transformer: Hierarchical Vision Transformer using Shifted Windows. Proceedings of the IEEE/CVF International Conference on Computer Vision, 10012–10022.\cite{swintransformer}
\\
(Completely Correct: Score=2, Similarity=0.995, Domain=Vision Transformer)\\
\midrule
Ma, X., \textbf{Xu, J., Sun, Y., He, Y., \& Lin, J. } (2022) Mega: Moving average equipped gated attention. \textbf{International Conference on Machine Learning, 162, 15369–15384}. \cite{mega} \\
 (Partially Hallucinated: Score=1, Similarity=0.798, Domain=Transformer)\\
\midrule
Kossen, J., Probst, P., Schirrmeister, R. T. \& Bischl, B. (2023) Self-Attention for Raw Numerical Tabular Data. IEEE Transactions on Neural Networks and Learning Systems. \\
(Completely Hallucinated: Score=0, Similarity=n.a., Domain=Tabular Transformer)\\
\bottomrule
   \end{tabular}
}
\end{table*}

\subsection{Experiment 1: Factual accuracy by domain}
\begin{figure*}[htb]
   \centering
   \includegraphics[width=0.98\linewidth]{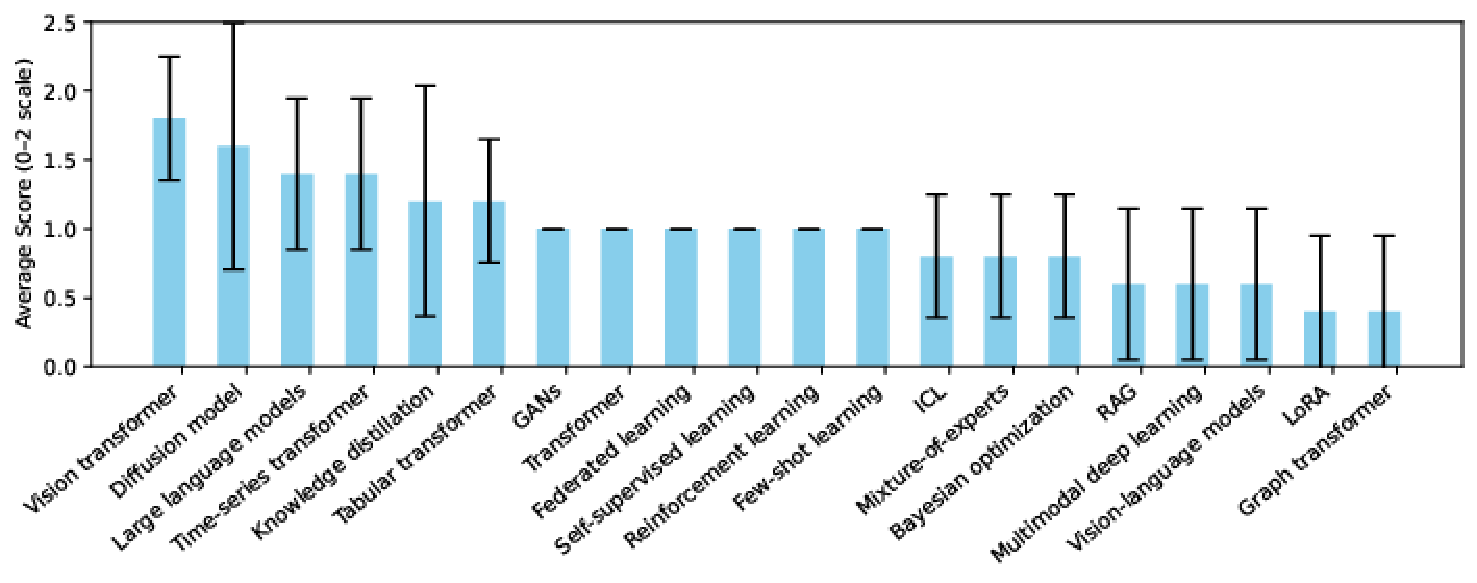}
   \caption{Mean factual accuracy score by research domain. Each domain contains five generated bibliographic records (score: 0 = completely hallucinated, 1 = partially hallucinated, 2 = completely correct). Error bars indicate the 95\% confidence interval.}\label{fig:domains}
\end{figure*}

As a preliminary analysis, we examine how factual accuracy varies across the twenty research domains. Fig.~\ref{fig:domains} presents the mean factual score for each domain. Overall, 15 out of 100 generated records were completely correct (score = 2), 68 were partially hallucinated (score = 1), and 17 were completely hallucinated (score = 0).

The accuracy varies considerably across domains. vision transformer and diffusion model achieved the highest accuracy, while LoRA and graph transformer exhibited the lowest scores. This variation is not simply explained by the complexity of metadata or the length of author lists. For example, Vision Transformer achieves the highest accuracy, yet as discussed in Experiment~4, even highly complex metadata such as the 12-author ViT paper \cite{vit} can be reproduced verbatim. Rather, the domain-level differences likely reflect the overall prominence of the research area in the pretraining corpus: domains with a concentration of highly cited landmark papers tend to achieve higher average accuracy, whereas domains that are either newer or more fragmented across many moderately cited works exhibit lower scores.

This observation motivates the subsequent experiments, which shift from domain-level aggregation to individual-level analysis. While domain-level accuracy provides a useful overview, it conflates papers with vastly different citation counts within the same domain. To disentangle the effect of citation frequency from domain membership, Experiments 2--4 analyze the relationship at the level of individual bibliographic records.

\subsection{Experiment 2: Citation frequency and factual accuracy} 
To test our central hypothesis that citation count serves as a proxy for training data redundancy, we first divided the 100 records at the median citation count ($\text{Mdn} = 818$) into low- and high-citation 
groups. 
A one-tailed t-test revealed that the high-citation group achieved significantly higher factual scores than the low-citation group: $p < .001$, $\text{Cohen's d} = 1.02$ ($M_\text{high} = 1.245$, $M_\text{low} = 0.725$).
This substantial difference, with the high-citation group scoring approximately 72\% higher on average and a large effect size, provides initial evidence that frequently cited papers are more accurately reproduced by the model.

\subsection{Experiment 3: Relationship between citations and fidelity} 
While the binary comparison in Experiment~2 provides initial evidence, it does not reveal the functional form of the relationship. Prior work has shown that memorization probability scales logarithmically with training data frequency \cite{carlini2}. We therefore hypothesize a log-linear relationship between citation counts and fidelity, and examine whether the effect is linear, logarithmic, or exhibits threshold behavior.

\begin{figure*}[thb]
   \centering
   \includegraphics[width=0.98\linewidth]{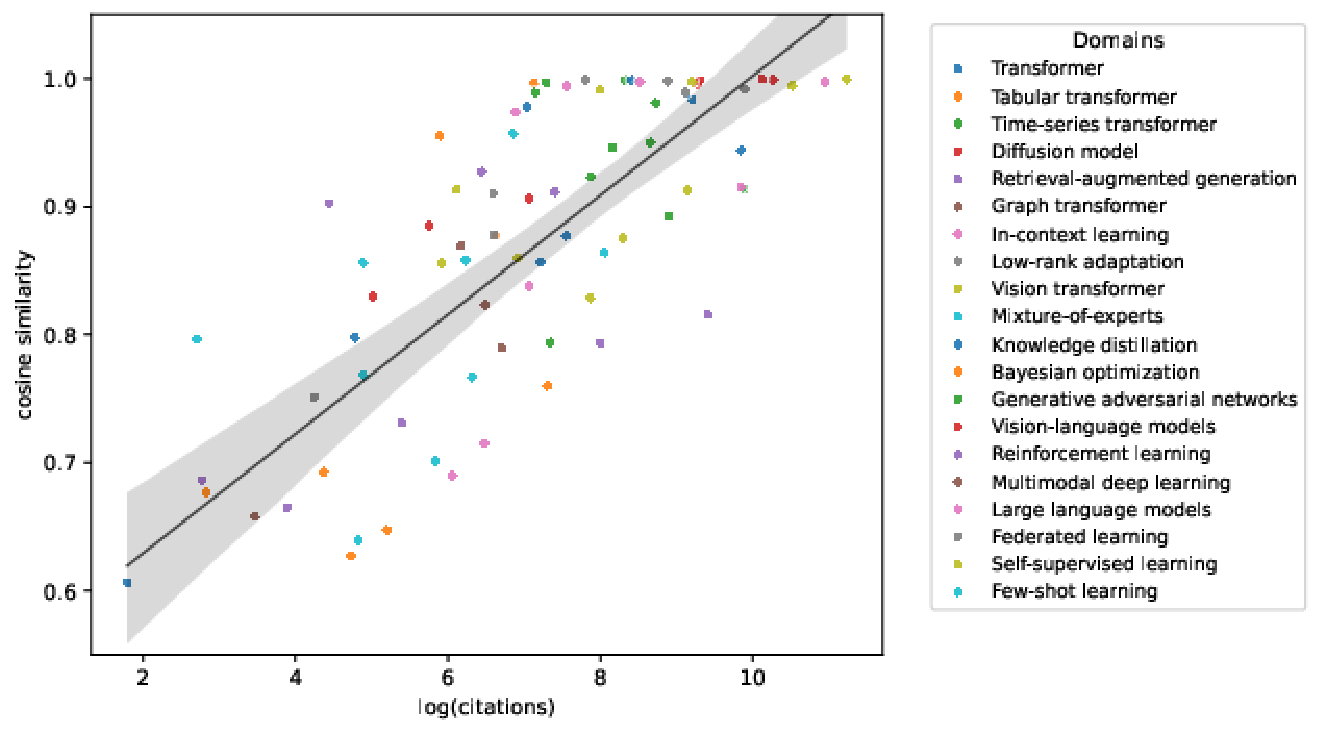}
   \caption{Relationship between citation frequency and generation fidelity \cite{niimi_icaart}. Each dot represents a factual record ($score > 0$), colored by research domain. The regression line indicates fitted linear regression with 95\% confidence interval (gray band). Strong correlation ($r = 0.75$, $p < .001$) demonstrates a log-linear scaling relationship. Note the saturation near $\log(\text{citation}) \approx 7$, suggesting a memorization threshold.}\label{fig:scatter}
\end{figure*}

Using only existing records ($\text{score} > 0$; $n = 81$), the visualized relationship between log-transformed citation counts and cosine similarity (Fig.~\ref{fig:scatter}) confirms a strong positive correlation ($r = 0.75$, $p < .001$) and linear regression indicates both intercept ($\text{coef} = 0.129$; $p < .043$) and slope ($\text{coef} = 0.088$; $p < .001$) statistically significant. 
The model explains 56\% of the variance ($R^2 = 0.56$), and the fitted regression line with 95\% CI demonstrates that this relationship holds consistently across domains.

In addition, residual analysis reveals systematic heteroscedasticity: the variance of prediction errors decreases markedly with citation count (Breusch-Pagan test: $\chi^2 = 7.11$, $p = .008$). 
Dividing the sample into tertiles by log(citation), residual variance in the high-citation group is approximately one-third that of the 
low-citation group ($var_{\text{low}} = 0.039$, $var_{\text{mid}} = 0.021$, $var_\text{high} = 0.012$). This pattern is visible in Fig.~\ref{fig:scatter} as tighter clustering near similarity = 1.0 at high citation counts, contrasted with substantial scatter at low citation counts.

We interpret this as evidence of a threshold behavior: highly cited papers appear repeatedly across diverse web sources, such as reference sections, educational materials, technical reports, and online discussions, leading to progressive memorization. The decreasing residual variance suggests that beyond a certain citation threshold, bibliographic information becomes nearly deterministic: the model transitions from probabilistic synthesis to verbatim recall.

\subsection{Experiment 4: Identifying memorization thresholds} 
Experiment 3 revealed a log-linear relationship between citations and fidelity, with patterns suggesting threshold behavior: high variance at low citations, diminishing to near-zero at high citations. To quantify these transitions, we model the non-linear pattern using logistic regression on min-max normalized cosine similarity.

The fitted model with the intercept ($\beta_0 = -2.360$, $p = .020$) and slope ($\beta_1 = 0.523$, $p = .003$) reveals two critical thresholds. First, the inflection point occurs at $-\beta_0 / \beta_1 \approx 4.51$, corresponding to approximately 90 citations. At this threshold, the model begins to transition from generative behavior (i.e., bibliographic information is synthesized from probabilistic token associations) to memorization-based behavior (i.e., specific records are increasingly recalled verbatim).

Second, we identify the saturation threshold (i.e., the minimum citation count at which near-perfect memorization consistently occurs). Unlike the inflection point, which is derived from the fitted logistic model, the saturation threshold is operationally defined from the onset of the near-ceiling region observed in the empirical data. As shown in Fig.~\ref{fig:scatter}, cosine similarity tends to cluster sharply near $1.0$ around $\log(\text{citation}) \approx 7$. In our sample, the minimum citation count in this saturation regime is 1,248. Table~\ref{tab:success} shows four representative examples from this regime, starting with the threshold case itself \cite{tabular_review}. Even at this boundary, all metadata fields are reproduced with near-perfect accuracy. Papers with substantially higher citations, such as ViT \cite{vit}, GPT-3 \cite{gpt3}, and Denoising Diffusion \cite{diffusion}, exhibit similarly perfect recall. Notably, even complex author lists (e.g., 12 authors for ViT) and detailed venue information are retained exactly, indicating that these records exist as discrete, retrievable units within the model rather than as distributed probabilistic representations.

\begin{table*}[tb]
\centering
   \caption{Representative records in the saturation area (cosine similarity
   $>$ 0.95) \cite{niimi_icaart}. The last entry is the identified saturation threshold (1,248 citations).}\label{tab:success}
   \begin{tabular}{
   m{16cm}
   }
\toprule 
\ccol{1}{\bfseries{Generated results}}\\
\midrule
Dosovitskiy, A., Beyer, L., Kolesnikov, A., Weissenborn, D., Zhai, X., Unterthiner, T., Dehghani, M., Minderer, M., Heigold, G., Gelly, S., Uszkoreit, J., \& Houlsby, N. (2021). An Image is Worth 16x16 Words: Transformers for Image Recognition at Scale. International Conference on Learning Representations.\cite{vit} \\
 (Citations=75,567, Similarity=0.999)\\
\midrule
Ho, J., Jain, A., \& Abbeel, P. (2020). Denoising Diffusion Probabilistic Models. Advances in Neural Information Processing Systems, 33, 6840–6851. \cite{diffusion} \\
 (Citations=28,944, Similarity=0.999)\\
\midrule
Brown, T. B., Mann, B., Ryder, N., Subbiah, M., Kaplan, J., Dhariwal, P., ... \& Amodei, D. (2020). Language Models are Few-Shot Learners. Advances in Neural Information Processing Systems, 33, 1877–1901. \cite{gpt3} \\
(Citations=56,858, Similarity=0.998)\\
\midrule
Gorishniy, Y., Rubachev, I., Khrulkov, V. \& Babenko, A. (2021). Revisiting Deep Learning Models for Tabular Data. Advances in Neural Information Processing Systems, 34, 18932–18943.\cite{tabular_review}\\
(Citations=1,248, Similarity=0.996)\\
\bottomrule
   \end{tabular}
\end{table*}

These thresholds suggest a two-stage memorization process. In the transition regime (90--1,248 citations), citation frequency increasingly determines accuracy, but substantial variance remains. Above the saturation threshold ($>$1,248 citations), papers are encoded nearly verbatim: the model has encountered these bibliographic records so frequently during pretraining that they are recalled deterministically, with minimal residual variance. This finding underscores that highly cited papers are not merely "well-represented" but are functionally memorized in a manner similar to the verbatim reproduction observed in privacy leakage studies \cite{carlini2}.

\section{Error Analysis}
While the saturation threshold identifies where verbatim recall begins, errors still occur even for highly cited papers. To understand these failure patterns, we analyze representative error cases across different citation ranges. Table~\ref{tab:error} shows four examples that reveal two distinct error mechanisms: hierarchical memorization, where metadata fields are recalled with varying fidelity depending on their frequency in the pretraining corpus, and memory interference, where semantically similar records are conflated.

\begin{table*}[thb]
\centering
   \caption{Representative error cases comparing generated (Gen.) and actual (Label) records \cite{niimi_icaart}. Fabricated elements are shown in bold. The first case illustrates memory interference among semantically similar papers (Section~5.2), while the remaining cases exhibit hierarchical memorization patterns (Section~5.1).}\label{tab:error}
   \begin{tabular}{
   c
   m{1.25cm}
   m{14cm}
   }
\toprule
\bccol{2}{Sample} & \bccol{1}{Results}\\
\midrule
&Gen. & Goodfellow, I., Pouget-Abadie, J., Mirza, M., Xu, B., Warde-Farley, D., Ozair, S., ... \& Bengio, Y. {\bfseries (2023).} Generative Adversarial {\bfseries Nets: A Review and New Perspectives}. Communications of the ACM, {\bfseries 66 (2), 84–96.} \\
\cmidrule{2-3}
&Label & Goodfellow, I., Pouget-Abadie, J., Mirza, M., Xu, B., Warde-Farley, D., Ozair, S., ... \& Bengio, Y. (2020). Generative adversarial networks. Communications of the ACM, 63(11), 139-144. \cite{gan} \\
\cmidrule{2-3}
&Metrics & Citations=19,612, Similarity=0.914\\
\midrule
& Gen. & Lewis, P., Perez, E., Piktus, A., Petroni, F., Karpukhin, V., Goyal, N., ... \& {\bfseries Riedel, S. (2023)}. Retrieval-Augmented Generation for Knowledge-Intensive NLP Tasks. {\bfseries Transactions of the Association for Computational Linguistics, 11, 945–963}.
 \\
\cmidrule{2-3}
&Label & Lewis, P., Perez, E., Piktus, A., Petroni, F., Karpukhin, V., Goyal, N., ... \& Kiela, D. (2020). Retrieval-augmented generation for knowledge-intensive NLP Tasks. Advances in neural information processing systems, 33, 9459-9474. \cite{rag}\\
\cmidrule{2-3}
&Metrics & Citations=12,218, Similarity=0.816\\
\midrule
& Gen. & {\bfseries Shazeer, N., He, Y., Strope, B., Roy, A., \& Wu, Y. (2023).} Switch Transformers: Scaling to Trillion Parameter Models with Simple and Efficient Sparsity. {\bfseries Advances in Neural Information Processing Systems, 36, 8580–8594.}
 \\
\cmidrule{2-3}
& Label & Fedus, W., Zoph, B., \& Shazeer, N. (2022). Switch transformers: Scaling to trillion parameter models with simple and efficient sparsity. Journal of Machine Learning Research, 23(120), 1-39.\\
\cmidrule{2-3}
&Metrics & Citations=3,143, Similarity=0.864\\
\midrule
& Gen. & {\bfseries Zhou, B., Li, X., Wu, Y., Yu, D., \& Yi, S. (2023)}. Self-Supervised Learning: Generative or Contrastive. {\bfseries Neural Networks, 163, 326–339.}\\
\cmidrule{2-3}
& Label & Liu, X., Zhang, F., Hou, Z., Mian, L., Wang, Z., Zhang, J., \& Tang, J. (2021). Self-supervised learning: Generative or contrastive. IEEE transactions on knowledge and data engineering, 35(1), 857-876.\cite{ssl}\\
\cmidrule{2-3}
&Metrics & Citations=2,613, Similarity=0.829\\
\bottomrule
   \end{tabular}
\end{table*}

\subsection{Hierarchical Memorization}
For high-citation papers ($>$10,000 citations), most author names are typically reproduced accurately, with errors primarily occurring in numeric information, such as volume, issue, and publication year. In contrast, low-citation papers ($<$5,000 citations) exhibit errors in all author names and their orders, suggesting incomplete memorization of the entire bibliographic record.

\begin{figure*}[htb]
   \begin{center}
   \includegraphics[width=\textwidth]{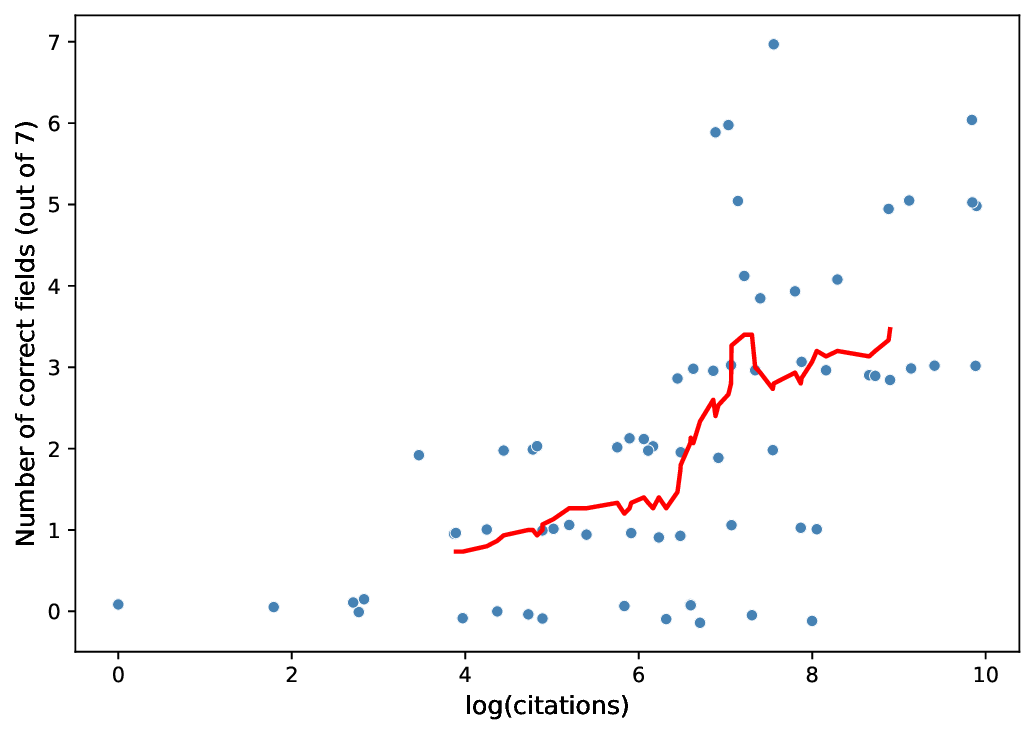}
   \caption{Number of correctly reproduced metadata fields (out of 7: title, first author, co-authors, year, journal, volume, pages) versus log-transformed citation count for partially hallucinated records (score $= 1$, $n = 68$). The red line indicates the moving average. Spearman $\rho = 0.667$, $p < .001$.}\label{fig:field_score}
   \end{center}
\end{figure*}

To quantify this pattern, we evaluated the correctness of seven metadata fields (title, first author, co-authors, year, journal, volume, and pages) for each partially hallucinated record (score $= 1$, $n = 68$). 
Field-level correctness was evaluated manually using normalized string matching. Case differences, punctuation differences, and common venue abbreviations (e.g., NeurIPS vs. Advances in Neural Information Processing Systems) were ignored. The first-author field was considered correct when the surname and initials matched. The co-author field was considered correct only when the generated co-author list matched the authentic list up to the level of detail provided by the model. Numeric fields such as year, volume, and pages were evaluated by exact match after normalization.
Figure~\ref{fig:field_score} shows the number of correctly reproduced fields as a function of log-transformed citation count. A strong positive correlation (Spearman $\rho = 0.667$, $p < .001$) confirms that higher-citation papers are reproduced with more complete metadata, rather than simply being ``correct or not.''

\begin{figure*}[p]
   \begin{center}
   \includegraphics[width=0.85\textwidth]{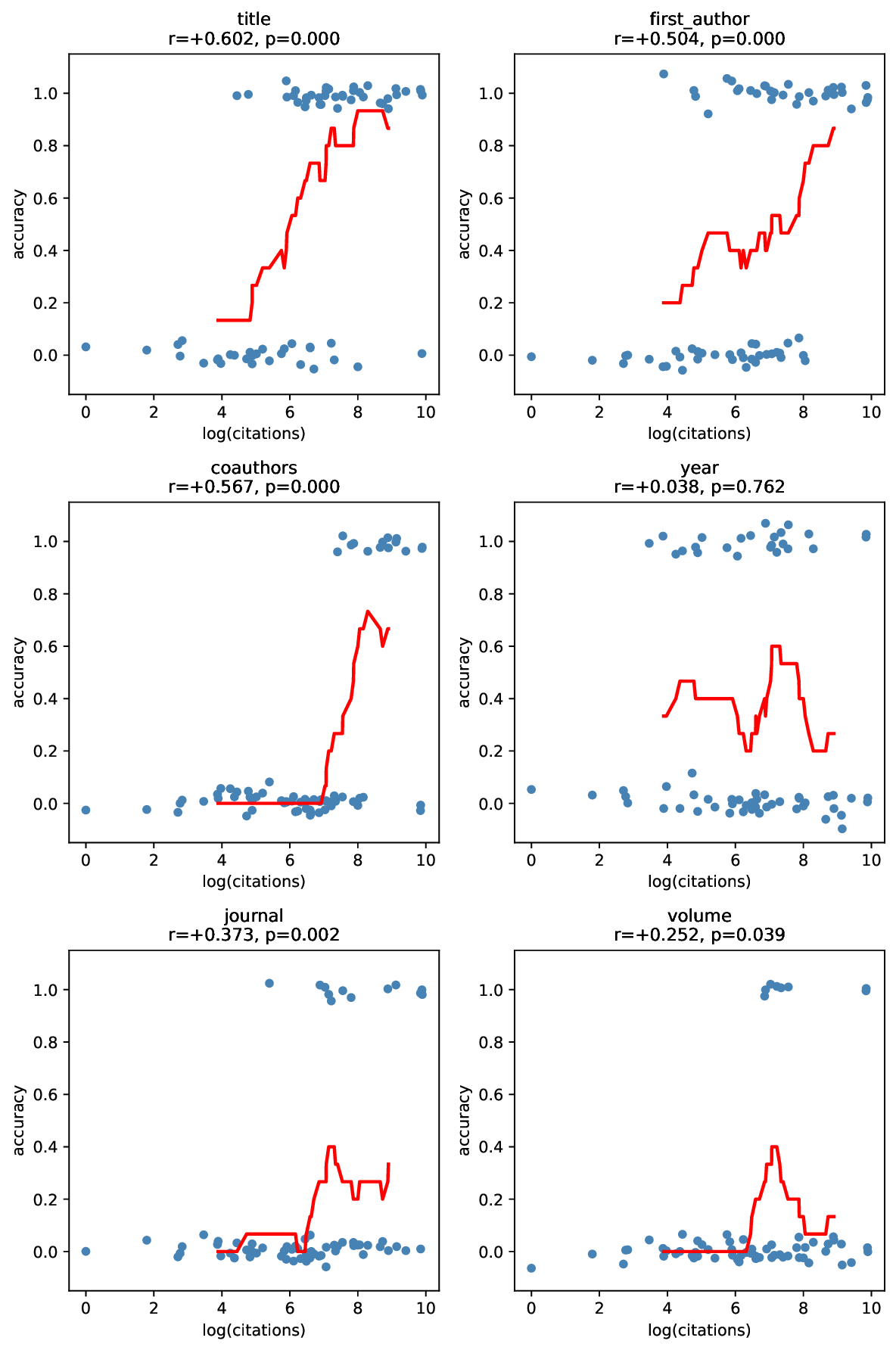}
   \caption{Field-level accuracy versus log-transformed citation count for partially hallucinated records (score $= 1$, $n = 68$). Each subplot shows the binary correctness (1 = correct, 0 = incorrect) of a single metadata field, with the red line indicating the moving average. Pearson $r$ and $p$-values are shown for each field.}\label{fig:by_field_accuracy}
   \end{center}
\end{figure*}

Furthermore, Fig.~\ref{fig:by_field_accuracy} breaks down accuracy by individual field, revealing distinct memorization trajectories. Title ($r = 0.602$) and first author ($r = 0.504$) show accuracy that increases steadily with citation count from relatively early on. In contrast, co-authors ($r = 0.567$) and journal name ($r = 0.373$) remain near zero for lower-citation papers and begin to rise only around $\ln(\text{citation}) \approx 7$ (i.e., $\sim$1,000 citations), suggesting that these fields require substantially greater training data redundancy before they are reliably memorized. Volume ($r = 0.252$) and pages ($r = 0.305$), while statistically significant, exhibit weaker correlations and remain noisy even at high citation counts. Notably, year shows essentially no correlation with citation count ($r = 0.038$, $p = .762$), standing apart from all other fields. We interpret this as evidence that the publication year is rarely retrieved as a memorized attribute of an individual record. Instead, the model appears to rely on a coarse domain-level prior: given the topical keywords of a query, it estimates a plausible publication period based on when research in that area was most active in the pretraining corpus, and outputs a year drawn from that distribution rather than from the bibliographic record itself. Under this view, even a paper whose title and first author are reproduced verbatim need not be accompanied by an accurate year, because the year is generated by a separate, distribution-level mechanism. Only for a small number of landmark papers, such as ViT \cite{vit}, is the year tightly co-memorized with the title, presumably because their publication dates appear redundantly in the pretraining corpus as part of the canonical reference itself. For the remaining records, exact year recall would require a degree of redundancy that even high citation counts do not appear to provide, which is consistent with the absence of any monotonic trend in the year subplot of Fig.~\ref{fig:by_field_accuracy}.

In general, author names, particularly the first author, and the title of the paper frequently appear in the pretraining corpus. Conversely, other information, such as co-author list, journal title, and numeric values (e.g., volumes, numbers, pages) are rarely detailed outside of the reference list in academic papers. For this reason, author names and paper titles are likely to be prioritized for memorization in LLMs. Furthermore, it is anticipated that the uneven distribution of information within pretraining corpora affects LLM hallucinations.

\subsection{Memory Interference}
The hierarchical memorization analysis above explains errors arising from insufficient training data redundancy: less-redundant fields are fabricated while more-redundant ones are retained. However, a different class of errors occurs even when citation counts are high. Rather than incomplete memorization, these errors stem from the conflation of multiple similar records.

A notable case is the highest-citation error record in Table~\ref{tab:error} ("Generative Adversarial Nets: A Review and New Perspectives," 19,612 citations, cosine similarity = 0.914). In this domain, multiple highly cited papers by the same lead author share nearly identical titles:
\begin{itemize}
\item Goodfellow et al. (2014) Generative adversarial nets. Advances in Neural Information Processing Systems. \cite{gan_original}
\item Goodfellow et al. (2020) Generative adversarial networks. Communications of the ACM. \cite{gan}
\end{itemize}
In addition, several survey papers on GANs with similar phrasing also exist \cite{gan1,gan2}. The generated record appears to conflate these sources: while the author list matches Goodfellow et al., the title incorporates survey-like phrasing ("A Review and New Perspectives") that does not belong to either original paper.

This suggests \textit{memory interference}: when multiple highly cited papers share similar titles and semantic content, the model conflates metadata across distinct bibliographic records, synthesizing a plausible but non-existent combination. This indicates that even for highly redundant knowledge, retrieval is not a simple lookup but rather a probabilistic reconstruction that can merge overlapping high-frequency patterns. In other words, unless a single bibliographic record appears frequently enough to override probabilistic token associations, it may not achieve accurate memorization.

This phenomenon can be interpreted through the lens of associative memory, which offers a useful conceptual framework for understanding the observed interference. In classical Hopfield networks \cite{hopfield1982}, learned patterns correspond to attractors in an energy landscape: given a partial or noisy input, the network converges to the nearest stored pattern. However, when multiple similar patterns are stored, their attractors can merge, giving rise to \textit{spurious attractors} that do not correspond to any individual stored pattern but rather to a blend of several. Ramsauer et al. \cite{ramsauer2021} have shown that the attention mechanism in Transformers can be reinterpreted as a modern Hopfield network, where the softmax-based attention weight computation corresponds to energy minimization in a continuous-state Hopfield model. Under this framework, memorized bibliographic records can be viewed as attractors, and the model's retrieval process as energy minimization over the stored patterns. When multiple highly cited papers share similar titles and authors, the corresponding attractors would lie close together in the energy landscape and may merge into a spurious attractor, causing the model to converge to a fabricated composite rather than any single authentic record. While this interpretation remains a conceptual analogy rather than a direct mechanistic proof, it offers a plausible explanation for the memory interference patterns we observe. This account also complements the empirical view of memorization as a form of overfitting to frequently encountered training sequences \cite{carlini2}: while overfitting explains \textit{why} certain records are memorized, the Hopfield framework suggests a possible mechanism for \textit{how} interference arises among similar memorized patterns.

\section{Conclusion}
\subsection{Key findings}
This study empirically examined how citation frequency functions as a proxy for hallucination in citation recommendation by LLMs. 
The model was instructed to output JSON-formatted results without explanations, effectively disabling
 IDK responses. In line with previous study \cite{openai_hallucinate}, such output constraints encourage the model to produce plausible yet non-existent entries. 

Our key findings are as follows:

\subsubsection{Strong correlation between citation count and factual accuracy.} High-citation papers achieve significantly higher factual accuracy than low-citation ones ($p < .001$, Cohen's $d = 1.02$). A log-linear relationship between citation count and cosine similarity ($r = 0.75$, $p < .001$, $R^2 = 0.56$) confirms that citation frequency serves as a reliable proxy for training data redundancy. Domain-level analysis further revealed that accuracy varies considerably across research areas, with domains containing highly cited landmark papers achieving higher average scores.

\subsubsection{Memorization thresholds.} Logistic regression identified two memorization thresholds: an inflection point at approximately 90 citations, where the model begins transitioning from probabilistic synthesis to memorization-based retrieval, and a saturation threshold at approximately 1,200 citations, beyond which bibliographic records are reproduced nearly verbatim with minimal residual variance.

\subsubsection{Hierarchical memorization.} Memorization does not operate uniformly across metadata fields. Field-level analysis of partially hallucinated records ($n = 68$) revealed that the number of correctly reproduced fields increases with citation count (Spearman $\rho = 0.667$, $p < .001$). Title ($r = 0.602$) and first author ($r = 0.504$) are memorized earliest, while co-authors and journal names begin to be accurately reproduced only beyond approximately 1,000 citations. Volume and page numbers remain unreliable even at high citation counts, and year shows no significant correlation with citation frequency.

\subsubsection{Memory interference.} Even for highly cited papers, errors can arise from the conflation of semantically similar records. When multiple papers share similar titles and authors, the model synthesizes a plausible combination from them. This phenomenon can be interpreted through the modern Hopfield network framework, where similar memorized patterns merge into spurious attractors.

\subsection{Implications}
Building on the view, established in our preliminary work \cite{niimi_icaart}, that hallucination and memorization arise from the same probabilistic learning process, the present study reframes that process as a graduated, hierarchically structured phenomenon rather than a clean dichotomy. Memorization is not a single state that a record either occupies or does not; it operates at two interacting levels of granularity. At the record level, fidelity scales log-linearly with citation frequency and crosses two empirically identifiable thresholds (an inflection around 90 citations and a saturation point around 1,200), so a model output can lie anywhere along a continuum from purely synthesized to verbatim recalled. At the field level, even within a single partially recalled record, individual metadata fields cross the memorization boundary at different citation levels, with title and first author memorized earliest and venue and numeric fields requiring substantially greater redundancy. The log-linear scaling, the two thresholds, the field-level memorization order, and the memory interference among similar high-frequency records are all consistent manifestations of this single underlying mechanism — the uneven, layered probability distribution of knowledge in the pretraining corpus.

While prior study \cite{carlini2} demonstrated that memorization emerges primarily when sufficient context is given, our results suggest the complementary mechanism: even with minimal prompting such as only specifying the domain keywords, highly redundant knowledge which is frequently represented in pretraining corpus can be recalled verbatim. In other words, context and redundancy are complementary for LLM memorization, and retrieving reasonably accurate information is possible even with only one of these elements specified.

Our findings further reveal a hierarchical structure in memorization. Paper titles and first author names are prioritized, while journal names and numeric values (volume, issue, pages) are more prone to fabrication. This hierarchy likely reflects uneven training data redundancy, as each element does not appear equally across the pretraining corpus. Author names and titles frequently appear in citations, acknowledgments, and web discussions, whereas venue details are typically confined to reference lists. For information with low redundancy, the model relies on probabilistic token associations rather than memorized sequences, resulting in plausible but fabricated metadata. Conversely, at high redundancy levels, verbatim recall emerges as a manifestation of overfitting, which occurs when probabilistic synthesis collapses into deterministic reproduction. Notably, even when citation counts exceed the saturation threshold, memory interference can still occur when multiple highly cited papers share similar titles and authors, leading the model to conflate their metadata.

Although this study focused on academic citation recommendation, the underlying mechanisms are not domain-specific. Any task that requires LLMs to reproduce structured factual records from pretraining knowledge (e.g., patent references, legal precedents, pharmaceutical data) is subject to the same redundancy-driven, hierarchically layered memorization dynamics, in which different fields of a single record can simultaneously sit at different points along the synthesis--memorization continuum. In these domains, the consequences of fabricated metadata may be far more severe, and the field-level memorization hierarchy and memory interference patterns identified here should inform how practitioners assess which parts of an LLM-generated output are likely to be reliably recalled and which are likely to be probabilistically reconstructed.

\subsection{Limitations}
This study has several limitations that should be addressed in future research.

\paragraph{Experimental scale.} Our analysis focused on a single model (GPT-4.1) with 100 bibliographic records (20 domains $\times$ 5 papers), constrained by the need for rigorous manual validation. The memorization thresholds identified here may differ across models with varying architectures, training data, and model sizes, and a larger sample could reveal more fine-grained patterns across subfields. That said, a study examining the effect of reasoning on bibliographic generation under strict constraints \cite{niimi_distortion} reported that fabrication of non-existent references persists across other frontier models (GPT-5.2 and Gemini 3 Flash) and at a larger scale (1,000 records), suggesting that hallucination in citation generation is not specific to a single model. Moreover, despite the limited sample size, statistically significant effects with large effect sizes were consistently observed (e.g., Cohen's $d = 1.02$, $r = 0.75$, Spearman $\rho = 0.667$), indicating sufficient statistical power to detect the core relationships. Nevertheless, systematic cross-model comparison of memorization thresholds remains an important direction for future work.

\paragraph{Domain scope.} To ensure comparable citation counts across papers, we limited our investigation to the computer science domain. Cross-disciplinary studies are needed to assess whether similar memorization thresholds exist when controlling for field-specific citation norms.

\paragraph{Language.} We evaluated only English-language publications. Multilingual contexts may present different memorization dynamics due to varying training data distributions across languages.

\paragraph{Temporal snapshot.} Citation counts were retrieved at a single time point (October 2025), but the relationship between citations and memorization may evolve as models are updated and retrained with newer corpora.

\paragraph{Publication types.} The ``academic papers'' in our experiments allow conference papers and preprints such as those on arXiv. While this broad scope is useful for exploring LLMs' internal memory and identifying memorization thresholds, actual user requests often involve stricter conditions (e.g., peer-reviewed journal articles only, or papers from specific time periods). Under such real-world constraints, LLMs may not be able to handle tasks merely by outputting memorized knowledge.

\paragraph{Reasoning capabilities.} Recent studies suggest that reasoning capabilities in LLMs, such as Chain-of-Thoughts (CoT) \cite{cot} and self-consistency \cite{llm_selfconsistency}, may improve task accuracy and reduce hallucinations \cite{reasoning_survey1,cot_effect1,cot_effect2,reasoning_survey2}. However, in a \textit{closed system} we focus on, where LLMs cannot rely on external tools, such reasoning may not necessarily resolve this limitation. A preliminary study has begun to examine this aspect \cite{niimi_distortion}, and further research is needed.

\subsection*{Acknowledgment}
This study was funded by JSPS KAKENHI (Grant No. JP\allowbreak 24\allowbreak K\allowbreak 16472)

\bibliographystyle{unsrt}
\bibliography{bibtex}

\end{document}